\def\year{2021}\relax

\documentclass[letterpaper]{article} 
\usepackage{aaai21}  
\usepackage{times}  
\usepackage{helvet} 
\usepackage{courier}  
\usepackage[hyphens]{url}  
\usepackage{graphicx} 
\usepackage{graphicx}  
\usepackage{natbib}  
\usepackage{caption} 
\usepackage{amsmath}
\usepackage{amssymb}
\usepackage{booktabs}
\usepackage{bm}
\usepackage{multirow}
\usepackage{graphicx}
\usepackage{subfigure}
\usepackage{xcolor}
\usepackage[switch]{lineno}
\usepackage{comment}

\title{
Modeling Heterogeneous Relations across Multiple Modes for Potential Crowd Flow Prediction
}

\author{Qiang Zhou\textsuperscript{\rm 1}, Jingjing Gu\textsuperscript{\rm 1,*}, Xinjiang Lu\textsuperscript{\rm 2}, Fuzhen Zhuang\textsuperscript{\rm 3,4}, Yanchao Zhao\textsuperscript{\rm 1}, Qiuhong Wang\textsuperscript{\rm 1}, Xiao Zhang\textsuperscript{\rm 5}\\
}
\affiliations{

    \textsuperscript{\rm 1}Nanjing University of Aeronautics and Astronautics,
    \textsuperscript{\rm 2}Business Intelligence Lab, Baidu Research,
    \textsuperscript{\rm 3}Key Lab of Intelligent Information Processing of Chinese Academy of Sciences (CAS), ICT, Beijing, China,
    \textsuperscript{\rm 4}University of Chinese Academy of Sciences, Beijing, China,
    \textsuperscript{\rm 5}Shandong University\\

    \{zhouqnuaacs, gujingjing\}@nuaa.edu.cn; luxinjiang@baidu.com; zhuangfuzhen@ict.ac.cn;\\ \{yczhao, baoh9491\}@nuaa.edu.cn;  xiaozhang@sdu.edu.cn

}

\begin{document}
\maketitle

\begin{abstract}

Potential crowd flow prediction for new planned transportation sites is a fundamental task for urban planners and administrators. Intuitively, the potential crowd flow of the new coming site can be implied by exploring the nearby sites. However, the transportation modes of nearby sites (e.g. bus stations, bicycle stations) might be different from the target site (e.g. subway station), which results in severe data scarcity issues. To this end, we propose a data driven approach, named MOHER, to predict the potential crowd flow in a certain mode for a new planned site. Specifically, we first identify the neighbor regions of the target site by examining the geographical proximity as well as the urban function similarity. Then, to aggregate these heterogeneous relations, we devise a cross-mode relational GCN, a novel relation-specific transformation model, which can learn not only the correlation but also the differences between different transportation modes. Afterward, we design an aggregator for inductive potential flow representation. Finally, an LTSM module is used for sequential flow prediction. Extensive experiments on real-world data sets demonstrate the superiority of the MOHER framework compared with the state-of-the-art algorithms. 

\end{abstract}

\section{Introduction}

Public transportation site selection acts as a pivotal part in urban planning. 
For example, is it appropriate to build a subway station in the target site/region when promoting urban infrastructure construction. 
To answer this question, existing methods, like \cite{lin2020multimoora, jelokhani2015decision}, mainly focus on how to balance the trade-offs between efficiency and cost, only a few studies~\cite{chen2015bike,aaai2020} focus on site selection through predicting future demands. 
How to fulfil the traffic demands in the future is the key to assess the quality of a site selection, especially for a transportation site.
Therefore, potential crowd flow prediction of new planned sites is one of the most fundamental and important tasks for the city planners and administrators.

\begin{figure}[t]
	\centering
	\subfigure[]{
 		\includegraphics[width=0.22\textwidth]{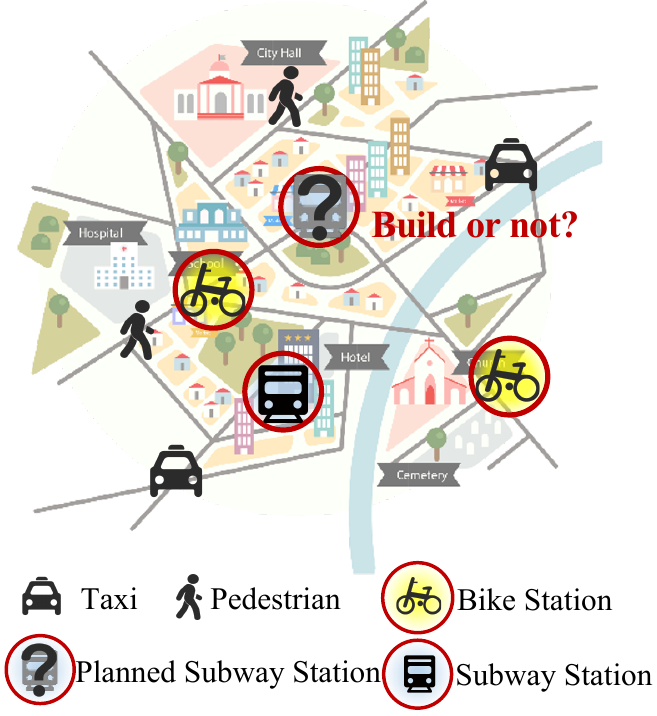}}
 	\subfigure[]{
 		\includegraphics[width=0.22\textwidth]{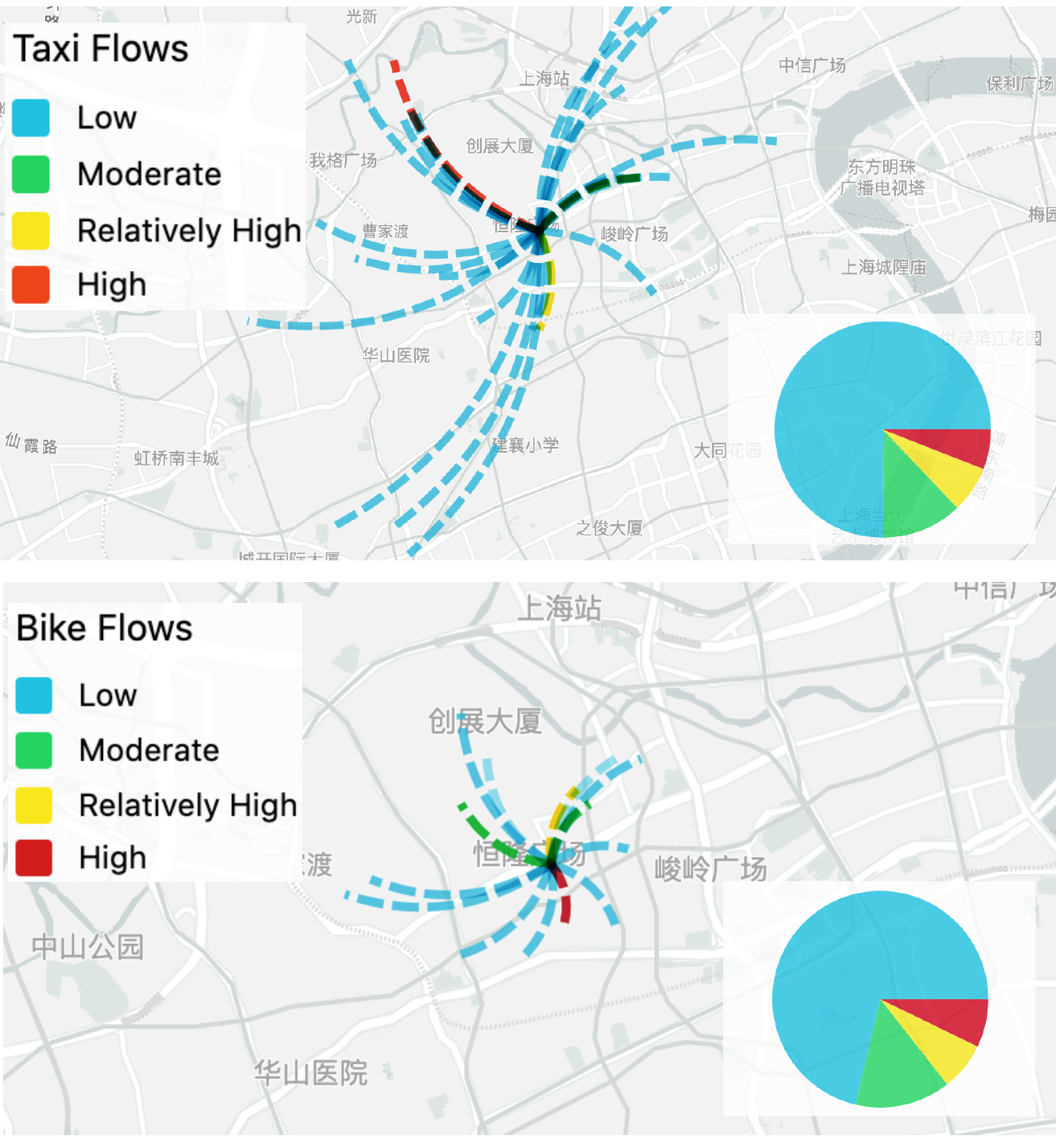}}
	\caption{(a): An example of the subway station placement problem. 
	(b): From the same origin, the outgoing crowd flows in different modes are different and similar: 1) different travel distances and destinations; and 2) similar traffic volume distributions depicted in the pie charts.}
	\label{fig:intro}
\end{figure}

In general, the potential crowd flow of a new planned site can be estimated by examining nearby sites in a collective way. 
For example, \cite{aaai2020} studies the potential flow prediction of a new subway station with neighbor subway stations. 
However, as the transportation sites in the same mode often locate far from each other, the transportation modes of nearby sites might be different from the planned target site as illustrated in Fig.~\ref{fig:intro}(a), which leads to severe data scarcity issues.

Recently, advanced matrix/tensor completion approaches, e.g.~\cite{babu2020sparse, mutinda2019time, takeuchi2017au}, have been proposed to solve the Kriging problems of unrecorded samples.
Some researchers~\cite{li2017citywide, xie2019active, chen2020nonconvex} have also put many efforts to solve the data sparsity problem in traffic completion and prediction. 
However, these methods are essentially transductive and could hardly be applied for new sites/nodes directly using trained models. 
Besides, it is not worthy to retrain the model from scratch only for a single site.

In this paper, we study the problem of potential crowd flow prediction for new planned transportation sites. 
Specifically, we propose a data-driven approach to MOdeling HEterogeneous Relations across multiple transportation modes for potential crowd flow prediction, named MOHER. 
We detail the MOHER framework in the following three aspects.

First, the relations between nearby sites in different transportation modes are highly dependent, dynamic and heterogeneous. 
As graph convolutional network (GCN) has shown its capability in characterizing complex dynamic spatio-temporal dependencies and the generalization potential to unseen nodes~\cite{hamilton2017inductive, nrgcn, yuan2020spatio}, 
we devise a cross-mode message passing network to model the cross-mode heterogeneous relations. 
We then leverage a neighbor aggregator to learn the inductive representations for each new site and further predict the potential crowd flows instead of retraining at every turn.

Second, the diversion phenomenon cannot be neglected when predicting the future crowd flow for a new planned site through the nearby sites. 
For instance, the new subway station will absorb a bulk of passengers who used to take buses. 
Besides, the new subway station will definitely attract more urban functions, which will compensate for the crowd flows of buses and bicycles. 
Along this line, we construct localized graphs centered by the target site/region to measure the geographical proximity and the functional similarity (characterized using surrounding Points Of Interest (POIs) for each category) simultaneously. 
Such that, the flow diversion and potential crowd flows can be captured automatically.

Third, 
apart from the diversion effect on the neighbor sites/regions, the planned new site will also have the diversion effect on other traffic modes \cite{liu2020incorporating},  where similarities and differences co-exist as illustrated in~Fig.~\ref{fig:intro}(b). 
Intuitively, contiguous sites should share similar travel patterns in terms of time and volume. 
On the other hand, due to the intrinsic characteristics (e.g. price and travel distance) of different transportation modes, the difference between crowd flows in different modes could be very large.
Thus, we introduce a novel relation-specific transformation model that can extract the similarity and difference of heterogeneous relations simultaneously.

To sum up, our contributions in this paper are as follows:

\begin{itemize}
\item 
We propose the inductive potential crowd flow prediction framework MOHER, which can naturally generalize to future new planning sites after training with only the historical flow data of the existing sites.
\item 
To capture the flow diversion and potential crowd flows of a new site, we respectively encode the geographical proximity and functional similarity among the nearby sites in different transportation modes.
\item 
We deeply explore the correlations and differences between cross-mode crowd flows and propose the Cross-Mode Relational GCN (CMR-GCN) to explicitly model each type of relations.
\item 
We conduct extensive experiments on real-world datasets, which demonstrate the effectiveness of MOHER for potential crowd flow prediction.
\end{itemize}

\section{Preliminaries}
We first introduce some important definitions and then formalize our problem. For brevity, we show notations and corresponding descriptions in Table \ref{notations}.

\begin{table}[ht]
\centering
\caption{Notations of symbols used in this work.}
\begin{tabular}{cl}
\toprule[1pt]
\textbf{Notations} & \textbf{Description} \\ \hline
$p \in \mathcal{P}$   & The transportation mode and its set. \\ \hline
\multirow{2}{*}{$v_{p} \in \mathcal{V}_p$}   & The site/region belonging to the mode \\ & $p$ and its set. \\ \hline
\multirow{2}{*}{$p_0, v_0$}  & The target transportation mode, \\ & the newly planned target site/region. \\ \hline
\multirow{2}{*}{$v^{\mathrm{Re}}, v ^{\mathrm{Un}} $}  & The region with/without historical \\ & crowd flow records. \\ \hline
\multirow{2}{*}{$\mathcal{X}_p^{(t)} = \{ x_{v_p}^{(t)} \}$} & The historical flow data of region $v_p$ \\ & at time $t$. \\ \hline
\multirow{2}{*}{$r_{\mathrm{GEO}}, r_{\mathrm{POI}}$} & The cross-mode edge type computed \\ & by the Geo-proximity/POI-similarity.   \\ \hline
\multirow{3}{*}{$\epsilon\!:\ (v_i, r, v_j)$} & The edge attribute between node \\ & (region/site) $i$ \& $j$ with edge type $r$. \\ & $r$ can be $r_{\mathrm{GEO}}$ or $r_{\mathrm{POI}}$.  \\ \bottomrule[0.5pt]
\end{tabular}
\label{notations}
\end{table}

\subsubsection{Definition 1}\textit{\textbf{Cross-mode Flow Relation Graph (CFRG).} A cross-mode flow relation graph centered by $v_0$ at time $t$ is a heterogeneous undirected weighted graph $G_{v_0}^{(t)}=(\mathcal{V}, \varepsilon, \mathcal{R})$, where $\mathcal{V}= \{ \mathcal{V}_p , \ p \in \mathcal{P} \}$ is a set of heterogeneous nodes (regions/sites) from all available transportation modes, $\varepsilon$ is the set of edge attributes of multiple relations, $\mathcal{R}= \{r_{\mathrm{GEO}} \} \cup \{r_{\mathrm{POI}}\}$ is the set of heterogeneous edge types. Hence, there may be more than one relations between two cross-mode nodes.}

\subsubsection{Definition 2}\textit{\textbf{Flow Prediction of New Planned Sites.} Assuming our new planned target region/site is $v_0$ with no data records belonging to the crowd flow $p_0$, we aimed at predicting the potential future flows $\bm{x}_{v_0}^{(t+1)}$ of the target $v_0$ with $p_0$. Therefore, the crowd flow prediction of new planned sites is formulated as a spatio-temporal prediction given a fixed temporal length $t'$:}

\begin{equation}
\begin{aligned}
\label{eqn:question}
\{ {(\mathcal{X}, G_{v_0})}^{(t-t'+1)},& {(\mathcal{X}, G_{v_0})}^{(t-t'+2)},\\ & \cdots, {(\mathcal{X}, G_{v_0})}^{(t)}\} \xrightarrow{f ( \cdot ) } x_{v_0}^{(t+1)}, 
\end{aligned}
\end{equation}

\begin{figure*}[t]
	\centering
	\includegraphics[width=\linewidth]{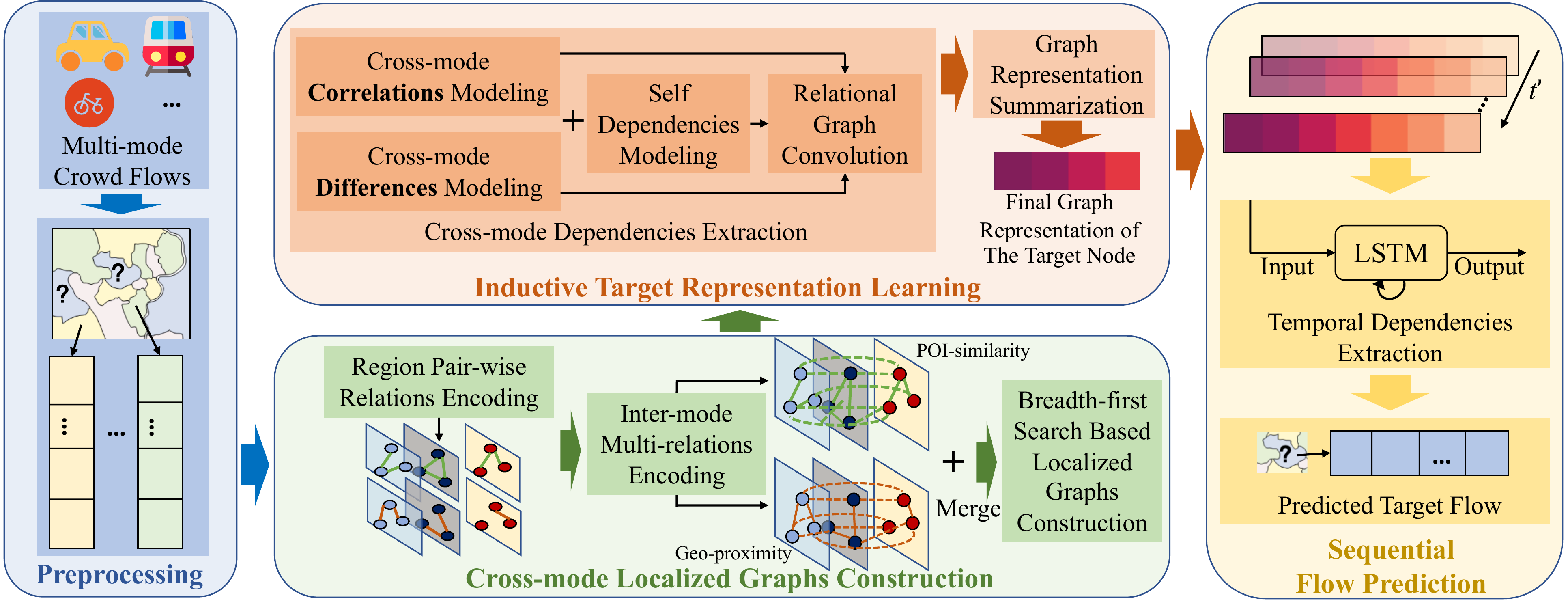}
	\caption{ An overview of the framework.}
	\label{fig:framework}
\end{figure*}
\noindent where $\mathcal{X} = \{ \mathcal{X}_p , \ p \in \mathcal{P}\}$ is the set of all cross-mode historical flow data. Eq. \ref{eqn:question} is to learn a function $f ( \cdot )$ that maps the $t'$ historical target-related cross-mode flow features to the target region flows in the next timestamp. 
To the best of our knowledge, we provide the first attempt on inductive potential crowd flow prediction of new planned sites with cross-mode flow data.

\section{Methodology}

In this section, we introduce our framework for inductive potential crowd flow prediction of new planned sites/regions by modeling heterogeneous relations across multiple transportation modes. 
The overview of MOHER framework is shown in Fig. \ref{fig:framework}, which consists of four major parts. 
(1) Pre-processing: collect a large number of multi-mode flow records and construct the input  $\mathcal{X}$. 
(2) Cross-mode Localized Graphs Construction: explore the potential crowd flows from nearby regions/sites via encoding both intra- and inter- mode relations according to the Geo-proximity and the POI-similarity, and construct localized graphs with the target node as the input. 
(3) Inductive Target Representation Learning: develop the cross-mode relational GCN for extracting the similarity and difference simultaneously between modes, then learn the inductive target node representation. 
(4) Sequential Flow Prediction: build an LSTM module to conduct sequential flow prediction of the target sites/regions.

\subsection{Cross-mode Localized Graphs Construction}

\subsubsection{Multiple pair-wise relations encoding}
Potential crowd flows are hidden in neighbor regions/sites of multiple transportation modes. However, aggregating inadequate information with only geographical distance could hardly characterize the flow features of new sites. The new construction of transportation sites can inspire the local functionality, which may cause incremental crowd flows of the new urban functional zone. Therefore, we encode two types of relations among sites/regions:
\begin{itemize}
\item \textbf{Geographical proximity.} Two regions/sites of multiple transportation modes near each other are affected by the same varying of travel demands. We choose geographic centers or station coordinates as the representatives of regions or sites, and then treat the normalized distance $\widetilde{\mathrm{dis}}_{ij}$ between arbitrary two regions $i$ and $j$ as inputs. The Geo-proximity is encoded as Eq. \ref{eqn:geo}.

\begin{equation}
\label{eqn:geo}
\epsilon_{ij,r_{\mathrm{GEO}}} = \left\{
\begin{aligned}
 &\mathrm{exp}(-\widetilde{\mathrm{dis}}_{ij}^2),&\mathrm{dis}_{ij} \le \gamma\\
& 0,&\mathrm{dis}_{ij} > \gamma
 \end{aligned}
\right.
\end{equation}
in which $\gamma$ is a threshold for the sparsity of $\epsilon_{r_{\mathrm{GEO}}}$.

\item \textbf{POI similarity.} Since POIs can indicate the region functionality, it is well instructive to refer to other regions with similar POI distribution for capturing the potential incremental crowd flows of new planned sites. In order to deal with the varying shapes and sizes of cross-mode nodes, we normalize the POI vectors $F$ of nodes and calculate the cosine distance between them by using Eq. \ref{eqn:poi}.

\begin{equation}
\label{eqn:poi}
\epsilon_{ij,r_{\mathrm{POI}}} = \left\{
\begin{aligned}
 &\mathcal{N}(\mathrm{Cos}(\widetilde{F_i}, \widetilde{F_j})),&\mathrm{Cos}(\widetilde{F_i}, \widetilde{F_j}) \ge \beta\\
& 0,&\mathrm{Cos}(\widetilde{F_i}, \widetilde{F_j}) < \beta
 \end{aligned}
\right.
\end{equation}
in which $\beta$ is a threshold for controlling the sparsity of $\epsilon_{r_{\mathrm{POI}}}$. After that, the results greater than zero are renormalized by $\mathcal{N}(\cdot)$ for more uniform values.
\end{itemize}

\subsubsection{Inter-mode relations encoding}
Potential crowd flows of new planned sites are not only involved in the target transportation mode, but we need to explore crowd flows from related regions/sites of multiple modes comprehensively. Therefore, we also encode the inter-mode relations by resorting to both \textbf{Geo-proximity} and \textbf{POI-similarity}. 

\begin{figure}[t]
	\centering
	\includegraphics[width=\linewidth]{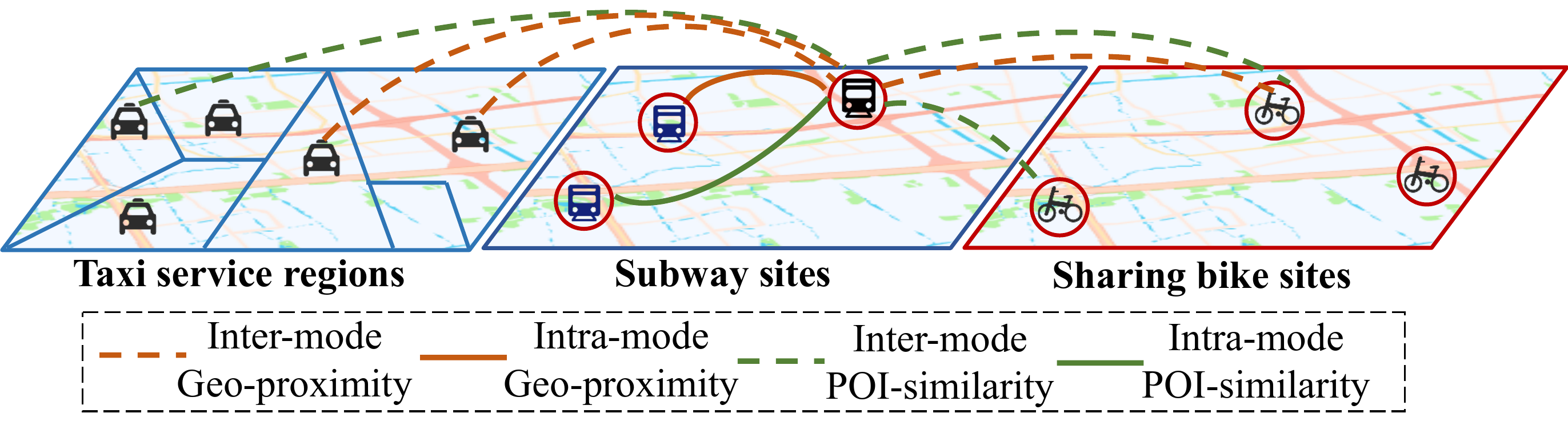}
	\caption{The diagrammatic sketch of establishing multiple heterogeneous relations for one region.}
	\label{fig:graph_cons}
\end{figure}

An example is shown in Fig. \ref{fig:graph_cons}. We establish multiple relations for one site/region with both intra-mode (e.g. subway stations) and inter-mode (e.g. bike stations/taxi service regions) sites/regions. Due to the diversities of cross-mode flow patterns, we consider each different relation across multiple transportation modes is heterogeneous.

Generally, we can model more types of relations with an extension of the above method. In this case, assuming we model $\rho$ ($\rho = 2$ in this work) types of relations among $|\mathcal{P}|$ transportation modes, there will be total $\frac{|\mathcal{P}| \times (|\mathcal{P}|+1) \times \rho}{2}$ types of heterogeneous edges. For cross-mode relations, the $\epsilon_{r_{\mathrm{GEO}}}$ and $\epsilon_{r_{\mathrm{POI}}}$ are larger as the overlapping area is larger, which means we can easily deal with the complex compositions of multiple transportation modes like Fig. \ref{fig:intro}.

\subsubsection{Breadth-first search based localized graphs construction}
Due to the continuous developments of urban infrastructure (e.g., new planned sites), the urban regional composition and functionality is evolving. Therefore, a static graph could hardly meet the demand of potential crowd flow prediction of new planned sites. In this paper, we extract $M$ neighbor nodes of the target region $v_0$ to construct localized graphs $\{ G_{v_0}^{(t-t'+1)},\cdots,G_{v_0}^{(t)}\}$ given timeslots $\{ t-t'+1, \cdots, t \}$ via breadth-first search. Particularly, the hyperparameter $M$ should be approximately equal to the size of the first and second-order neighborhood of target regions for involving most of the related information \cite{hamilton2017inductive}. As a result, our method can focus on the localized information and apply to the evolving graph structure for inductive regional flow prediction.

\begin{table}[t]
\centering
\begin{tabular}{l}
\toprule[1pt]
\textbf{Algorithm 1} Breadth-first search based localized graph\\ 
construction for the target node \\ \hline
1: \textbf{Input:} $M$, the target node $v_0$, the historical recorded\\
\quad  nodes $v^{\mathrm{Re}}$, the timeslot $t$\\
2: \textbf{Output:} the $(M+1)$-nodes localized graph $G_{v_0}^{(t)}$\\
3: \textbf{Init} size = 0, seq = $\{ v_0 \}$;\\
4: \textbf{while} size$<M+1$ and seq!=\{\} \textbf{do}\\
5:  \quad $v_i$ = seq.dequeue();\\
6:  \quad add $v_i$ into $G_{v_0}^{(t)}$, add relations of $v_i$ into $G_{v_0}^{(t)}$;\\
7:  \quad size = size + 1;\\
8:  \quad \textbf{for} $v_j$ in $v^{\mathrm{Re}}$ \textbf{do}\\
9:  \qquad   calculate $\epsilon_{ij,r_{\mathrm{GEO}}}$ with Eq.\ref{eqn:geo};\\
10: \ \ \textbf{end for}\\
11: \ \ enqueue $v_j$ into seq by the descending order\\
\qquad of $\epsilon_{ij,r_{\mathrm{GEO}}}$ with $\epsilon_{ij,r_{\mathrm{GEO}}}>0$, $v_j \notin G_{v_0}^{(t)} \cup$ seq;\\
12: \quad \textbf{for} $v_j$ in $v^{\mathrm{Re}}$ \textbf{do}\\
13: \qquad   calculate $\epsilon_{ij,r_{\mathrm{POI}}}$ with Eq.\ref{eqn:poi};\\
14: \ \ \textbf{end for}\\
15: \ \ enqueue $v_j$ into seq by the descending order\\
\qquad of $\epsilon_{ij,r_{\mathrm{POI}}}$ with $\epsilon_{ij,r_{\mathrm{POI}}}>0$, $v_j \notin G_{v_0}^{(t)} \cup$ seq;\\
16:\textbf{end while}\\
17:\textbf{return} $G_{v_0}^{(t)}$;
\\ \bottomrule[0.5pt]
\end{tabular}
\label{subgraph}
\end{table}

\subsection{Inductive Target Representation Learning}

\subsubsection{Cross-mode dependencies extraction} Given heterogeneous graphs constructed by Algorithm 1, we should explicitly model each kind of the heterogeneous cross-mode relations to extract flow embeddings and then map the localized graphs information into the potential flow patterns of the new planned site. Accordingly, we propose the Cross-Mode Relational Graph Convolutional Network (CMR-GCN) as the message passing module for modeling \textbf{correlations} and \textbf{differences} of heterogeneous relations simultaneously. 

Specifically, for each node $v_i, 1\leq i\leq M $ in a localized graph except the unrecorded target node, we extract the regional \textbf{correlations} $\{ \mathcal{C}_{i,r}^{l+1} | r \in \mathcal{R}\}$ at layer $l+1$ of each kind of heterogeneous relations via (4):

\begin{equation}
\label{eqn:corr}
\mathcal{C}_{i,r}^{l+1} =\sum_{j \in \mathcal{N}_{r}(i)}\sigma(\frac{\varepsilon_{r,ij}}{|\mathcal{N}_{r}(i)|} \bm{x}_j^l \bm{W}_{r,c}^l +\bm{b}_{r,c}^l ),
\end{equation}

\noindent where $\bm{x}_j^l$ denotes the features of node $j, 1\leq j\leq M$ at layer $l$, $\{ \bm{W}_{r,c}^l | r \in \mathcal{R}\}$ and $\{ \bm{b}_{r,c}^l | r \in \mathcal{R}\}$ are learnable parameters, $\mathcal{N}_{r}(i)$ denotes the set of $i$'s neighbors that connect to $i$ with relation category $r$. $\sigma$ denotes the ReLU activation function. 

In addition to the correlations, there are definite differences between cross-mode regional flow features. If the information of differences is ignored, there may even produce a negative transfer effect. Hence, we also model the \textbf{differences} $\{ \mathcal{D}_{i,r}^{l+1} | r \in \mathcal{R}\}$ between cross-mode node features:

\begin{equation}
\label{eqn:diff}
\mathcal{D}_{i,r}^{l+1} =\sum_{j \in \mathcal{N}_{r}(i)}\omega(\frac{\varepsilon_{r,ij}}{|\mathcal{N}_{r}(i)|} |\bm{x}_j^l-\bm{x}_i^l| \bm{W}_{r,d}^l +\bm{b}_{r,d}^l ),
\end{equation}

\noindent where $|\bm{x}_j^l-\bm{x}_i^l|$ represents the natural gap between node $i$ and neighbor node $j$, and $\omega$ is set as $tanh$. An intuitive explanation of Eq. \ref{eqn:diff} is that we construct a shadow relation $r_s$ for each kind of relation $r$ to indicate the differences between two original nodes connected by $r$. Consequently, we aggregate the shadow node $j_s$'s information which is D-value between node features into $\mathcal{D}_{i,r}^{l+1}$ via $r_s$. The correlations and differences of each kind of heterogeneous relations are accumulated on the transformed self-features, then the forward-pass-update formula of $\bm{x}_i^{l+1}$ is given by:

\begin{equation}
\label{eqn:acc}
\bm{x}_i^{l+1} =\sigma( \bm{x}_i^l \bm{W}_p^l + \sum_{r\in \mathcal{R}}(\mathcal{C}_{i,r}^{l+1}+\mathcal{D}_{i,r}^{l+1})).
\end{equation}

Finally, assuming there are $L$ message passing layers CMR-GCNs, the final node representation $\bm{g}_i$ of node $v_i$ is concatenated by the outputs of layer 1 to layer $L$:

\begin{equation}
\label{eqn:noderep}
\bm{g}_i = \mathrm{concat}(\bm{x}_i^{1}, \bm{x}_i^{2},\cdots  ,\bm{x}_i^{L})
\end{equation}

\subsubsection{Regularization for modeling heterogeneous relations} 
With both correlations and differences modeling, the total types of heterogeneous relations will be double to $|\mathcal{P}| \times (|\mathcal{P}|+1) \times \rho$ plus the number of shadow relations $\{ r_s | r \in \mathcal{R} \}$, which will lead to a surge in the number of parameters at the same time. To reduce the complexity of our model in case of overfitting, we apply matrix reconstruction to each of the learnable parameters:

\begin{equation}
\label{eqn:regularize}
\begin{split}
\bm{W}_{r,c}^l &= \sum_{q=1}^{Q}a_{\bm{W}_{r,c},q}^{l}B_{\bm{W},q}^{l}, \ 
\bm{W}_{r,d}^l = \sum_{q=1}^{Q}a_{\bm{W}_{r,d},q}^{l}B_{\bm{W},q}^{l} \\
\bm{b}_{r,c}^l &= \sum_{q=1}^{Q}a_{\bm{b}_{r,c},q}^{l}B_{\bm{b},q}^{l}, \ 
\bm{b}_{r,d}^l = \sum_{q=1}^{Q}a_{\bm{b}_{r,d},q}^{l}B_{\bm{b},q}^{l}
\end{split}
\end{equation}

Accordingly, each learnable parameter is reconstructed by basis $B$ and corresponding coefficients $a$. Along such a reconstruction process, all the parameters that model the correlations and differences share the same basis matrices respectively, which can also mitigate overfitting when some kinds of relations are uncommon in the training set.

\begin{figure}[t]
	\centering
	\includegraphics[width=\linewidth]{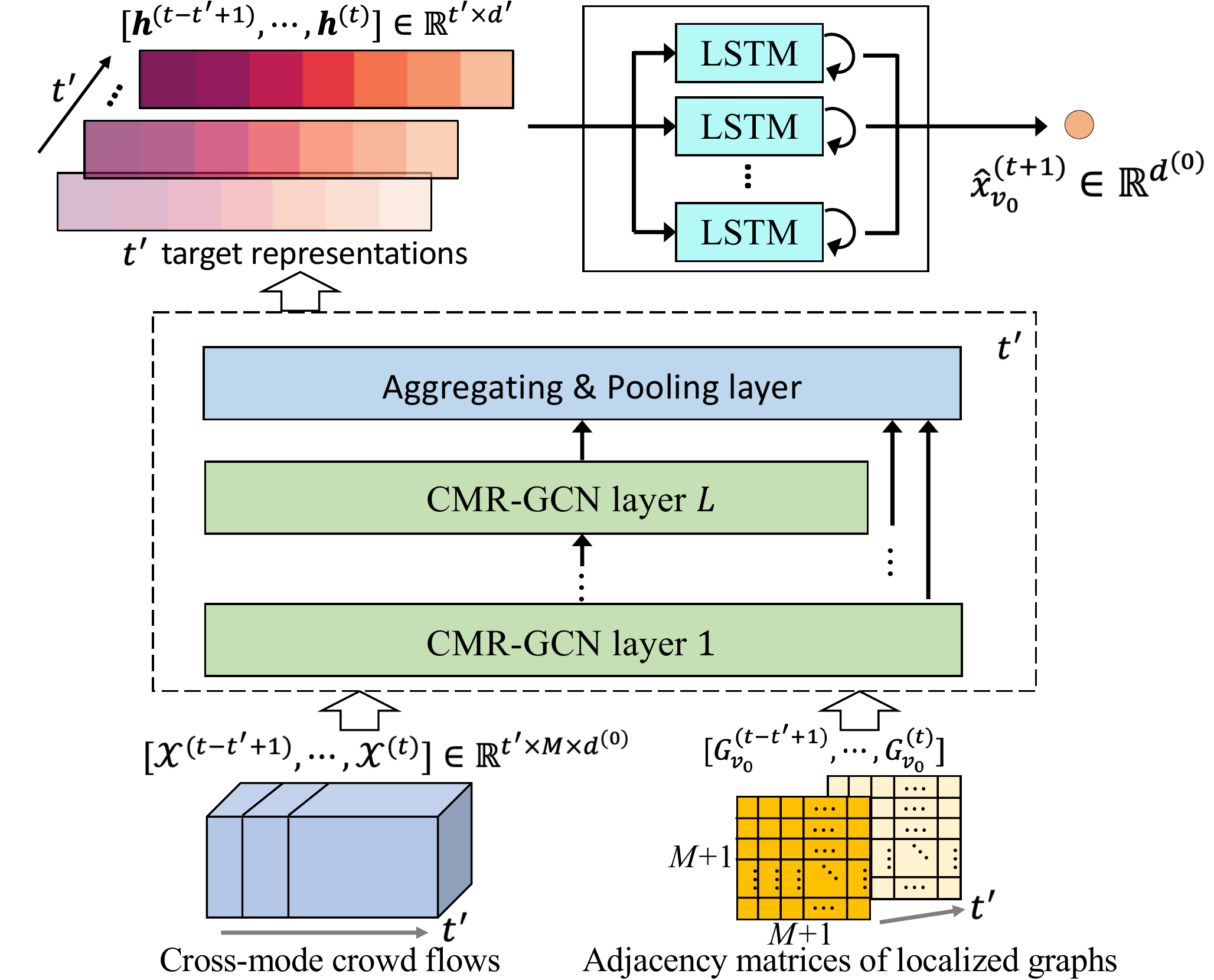}
	\caption{The network structure of our MOHER, where $d^{(0)}$ and $d'$ are the dimensions of origin and final flow representations. The shared network across different timeslots in the dotted box is applied to learn inductive cross-mode target representation.}
	\label{fig:network}
\end{figure}

\subsubsection{Graph representation summarizing} After the $L$ layers of regularized CMR-GCNs, each node $v_i$ except the target $v_0$ gets its flow embedding $\bm{g}_i$. Inspired by max-pooling operation, we propose a nearest neighbor pooling aggregator that selects the most related nodes connected to target $v_0$ by each existing type of relations, which can directly reduce the redundancy of features. Thus, the representation of target $v_0$ is constructed as:

\begin{equation}
\label{eqn:targetrep}
\bm{h} = \mathrm{concat}(\varepsilon_{v_0,r}\cdot \bm{g}_r | r \in \mathcal{R}_{v_0}),
\end{equation}

\noindent where $\mathcal{R}_{v_0}$ is a subset of $\mathcal{R}$ which contains the relations connected to $v_0$, $\bm{g}_r$ is the nearest neighbor of $v_0$ connected by relation $r$, and $\varepsilon_{v_0,r}$ is the value of the edge. Here $\bm{h}$ is the flow representation of the target node $v_0$. Finally, we adopt LSTM networks to extract temporal dependencies for predicting the $t+1$ target regional or station-level flows. The network structure is shown in Fig. \ref{fig:network}. Note that our proposed MOHER is inductive, it can generalize to future planned sites along with the development of the city.

\subsection{Model Training}
In the training phase, our model is optimized by minimizing the loss of mean square error (MSE) using Adam optimizer.

It is difficult to retrieve the historical data before the construction of existed sites. Therefore, we propose a simulation strategy to approximate real historical flows. Specifically, we randomly select existed sites to simulate planned sites, and we distribute the historical flow of them to the first-order Geo-proximate neighbors according to the product of the flow proportions and the connected edge weights. These selected sites are used as training targets.

\section{Experiments}

\subsection{Experiment Settings}

\subsubsection{Datasets} 
We construct a benchmark with three real-world flow datasets collected from NYC OpenData. Two of them contain the trip records of the taxi and bike in NYC. The other contains the turnstile usage condition of each subway station in NYC. All of them are generated in the time period from January 1st, 2018 to December 1st, 2019 (700 days).
\begin{itemize}
\item \textbf{NYC Turnstile Usage of Subway Stations:} NYC Metropolitan Transportation Authority provides the turnstile usage counts of NYC subway stations. There are about 5042 turnstiles used as entry/exit registers at 425 subway stations. The value records per 4 hours consist of following information: station-ID, turnstile-ID, record time, Entries value, Exits value, etc.
\item \textbf{NYC Citi Bike:} NYC Bike Sharing System generates the Citi Bike orders including 38 million and 100 thousand transaction records of 647 stations in the selected time period. This data set contains following information: bike pick-up station, bike drop-off station, bike pick-up time, bike drop-off time, trip duration.
\item \textbf{NYC Taxi:} NYC Taxi consists of about 187 million taxicab trip records. On average, there are about 267 thousand records in 262 regions every day. The records consist of following information: pick-up time, drop-off time, pick-up region, drop-off region, trip distance, etc.
\end{itemize}

\subsubsection{Baselines} We compare our framework with the following methods:
\begin{itemize}
\item \textbf{NA-HA:} Neighbor Average and Historical Average \cite{kamarianakis2003forecasting}. We model the spatial similarity and use the average of neighbor sites of the same transportation modes as the historical flows. Then, we predict the target crowd flow using the mean value over time slots.
\item \textbf{MGCN-MLP:} From Multiple Graph Convolutional Network \cite{geng2019spatiotemporal}, we aggregate information for modeling spatial dependency by multiple graphs. 
Then, a Multi-Layer Perceptron is used for mapping graph representation to the prediction result.
\item \textbf{NA-LSTM:} The feature representation of historical flows is similar to NA-HA, then the Long-Short-term-Memory method \cite{lstm} is introduced for temporal dependency modeling. 
\item \textbf{LP-GLP:} We use Label Propagation to complete the historical flows and construct a semi-supervised learning problem on graphs. Generalized Label Propagation \cite{GLP} is a semi-supervised learning method that produces flow representations of different times by exploring graph relations. Then, a Feedforward Neural Network (FNN) is used for prediction.
\item \textbf{STMGCN:} STMGCN leverages complete convolutional structures to model both temporal and multiple spatial dependencies. MGCN is applied in the ST-Conv blocks \cite{stgcn} for modeling multiple relations.
\item \textbf{MLC-PPF:} The Multi-view Localized Correlation learning model\cite{aaai2020} for subway Potential Passenger Flow uses the economic statistics as cross-domain knowledge, which just uses one mode flow.
\end{itemize}

\subsubsection{Metrics} We evaluate performances based on two popular metrics, which are Root Mean Square Error (RMSE) and Mean Absolute Percentage Error (MAPE) \cite{geng2019spatiotemporal, stgcn, ye2019co, deng2016latent}.

\subsubsection{Implementations} We aggregate crowd flows into 4 hours windows and apply Z-score normalization. We randomly selected 20\% sites for validation and 10\% for testing. They are deleted from the existing sites, and we construct the train set based on the remained 70\% site with flow data of the previous 60\% time (420 days). The validation and test set are constructed based on the respectively selected sites with flow data of the following 40\% time (280 days). We tune the hyperparameters using grid search based on the performance on the validation set, and report the average performances on the test set over 10 runs. Due to the unimode flow, NA-HA, LP-GLP, and MLC-PPF use the data just from the target mode. And we use 1 layer of MGCN and 2 layers of MLP with 64 hidden units and a dropout rate of 0.2 for MGCN-MLP. The modules in STMGCN is the same with MGCN-MLP but it is extended with Gated Linear Units of $K_t=t'/2$ and $K_t=t'/2+1$. The statistics used in MLC-PPF are the number of various kinds of commercial POIs. Other hyperparameters of baselines are the same as the grid searched hyperparameters of our method. All experiments are run on a PC with an NVIDIA GeForce RTX 2080Ti GPU, and 64 GB RAM running the 64-bit Ubuntu 16.04 system, python3.6, TensorFlow 2.0.

\subsection{Experimental Results}
Table \ref{results} shows the test errors comparison of different methods for potential crowd flow prediction of new planned sites.

\subsubsection{Results summary}
We can observe: (1) the models which can synthesis both spatial and temporal dependencies (e.g., LP-GLP, STMGCN, and proposed MOHER) achieve better performances. The proposed MOHER which can sufficiently model each type of cross-mode relations outperforms other baselines. Compared with the second-best result, MOHER decreases the errors by (9.1\%, 6.8\%) under RMSE and (37.0\%, 13.3\%) under MAPE. (2) The methods which model the multiple relations perform better than NA-HA or NA-LSTM. Meanwhile, the methods which especially consider the temporal dependency achieve better performances than NA-HA or MGCN-MLP, especially on the NYC Citi Bike Dataset. (3) MLC-PPF performs worse than our MOHER for the prediction of the long-term evolving targets due to the lack of cross-mode information and spatial dependency modeling. As the sharing bike is often used as an alternative transportation mode, the influence of cross-mode crowd flows on bikes is much more important than that on the subway, which may result in the worse performances of MLC-PPF on the bike dataset.

\begin{table}[t]
\centering
\caption{Potential Crowd flow forecasting error of new planned sites given by RMSE and MAPE on NYC Subway Turnstile Usage Dataset and NYC Citi Bike Dataset.}
\begin{tabular}{ccccc}
\toprule[1pt]
\multirow{2}{*}{Methods}         & \multicolumn{2}{c}{Subway} & \multicolumn{2}{c}{Bike} \\
                                 & RMSE  & MAPE            & RMSE                & MAPE               \\ \hline
NA-HA                            & 2182.2                & 3.942           & 20.967              & 1.828              \\
MGCN-MLP                         & 2060.6                & 3.185           & 23.020              & 1.528              \\
NA-LSTM                          & 1947.1                & 1.955           & 18.177              & 1.015              \\
LP-GLP                           & 1986.0                & 1.799           & 19.021              & 0.826              \\
STMGCN                           & 1821.2                & 1.784           & 19.230              & 0.988              \\
MLC-PPF                          & 1757.5                & 1.931           & 22.612              & 1.532              \\
\textbf{MOHER}                 & \textbf{1597.3}       & \textbf{1.124}  & \textbf{16.767}     & \textbf{0.716}     \\ \bottomrule[0.5pt]
\end{tabular}
\label{results}
\end{table}

\subsubsection{Case Study} 
\begin{figure}[t]
	\centering
	\subfigure[]{
 		\includegraphics[width=0.48\linewidth]{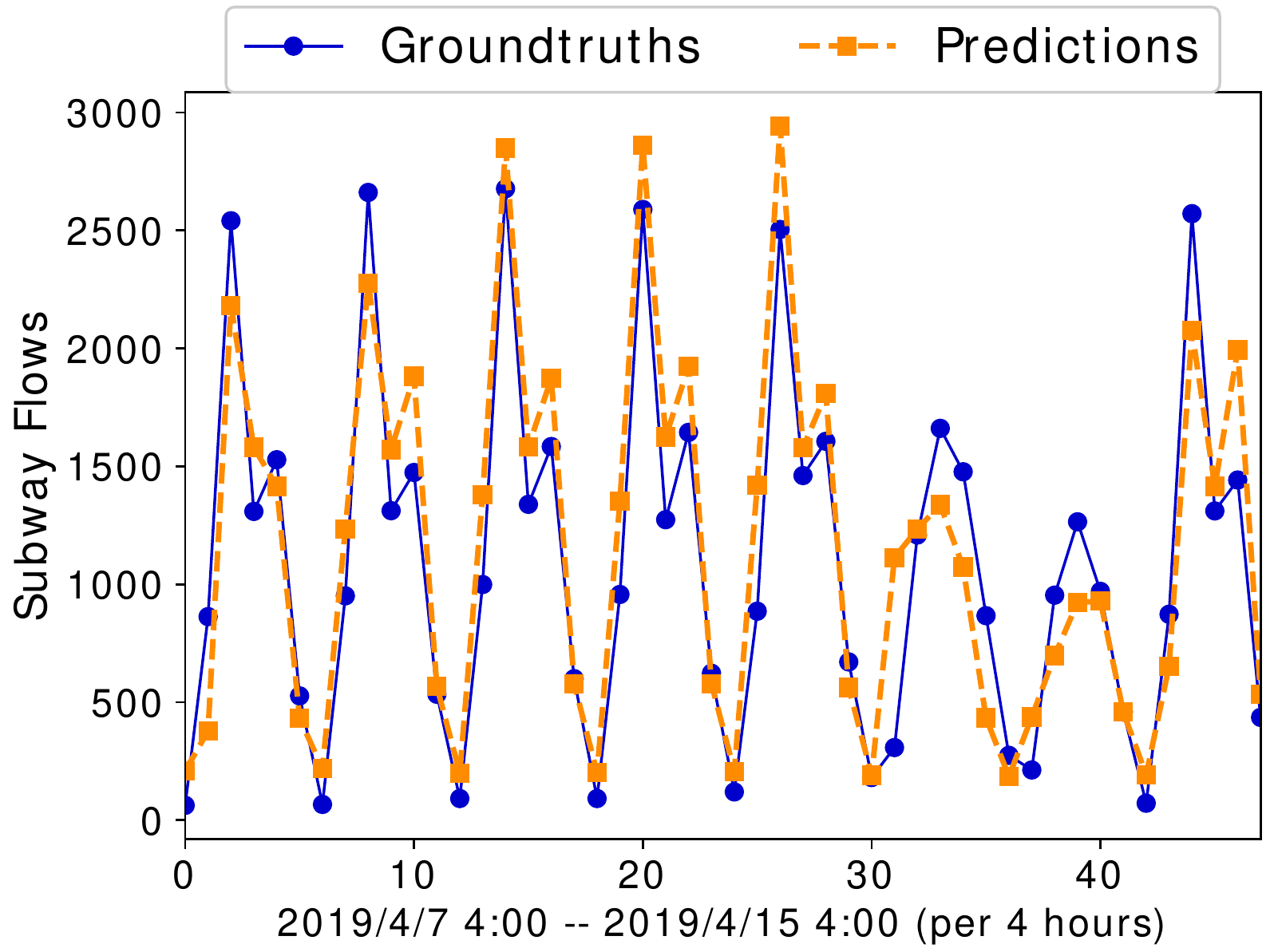}}
 	\subfigure[]{
 		\includegraphics[width=0.48\linewidth]{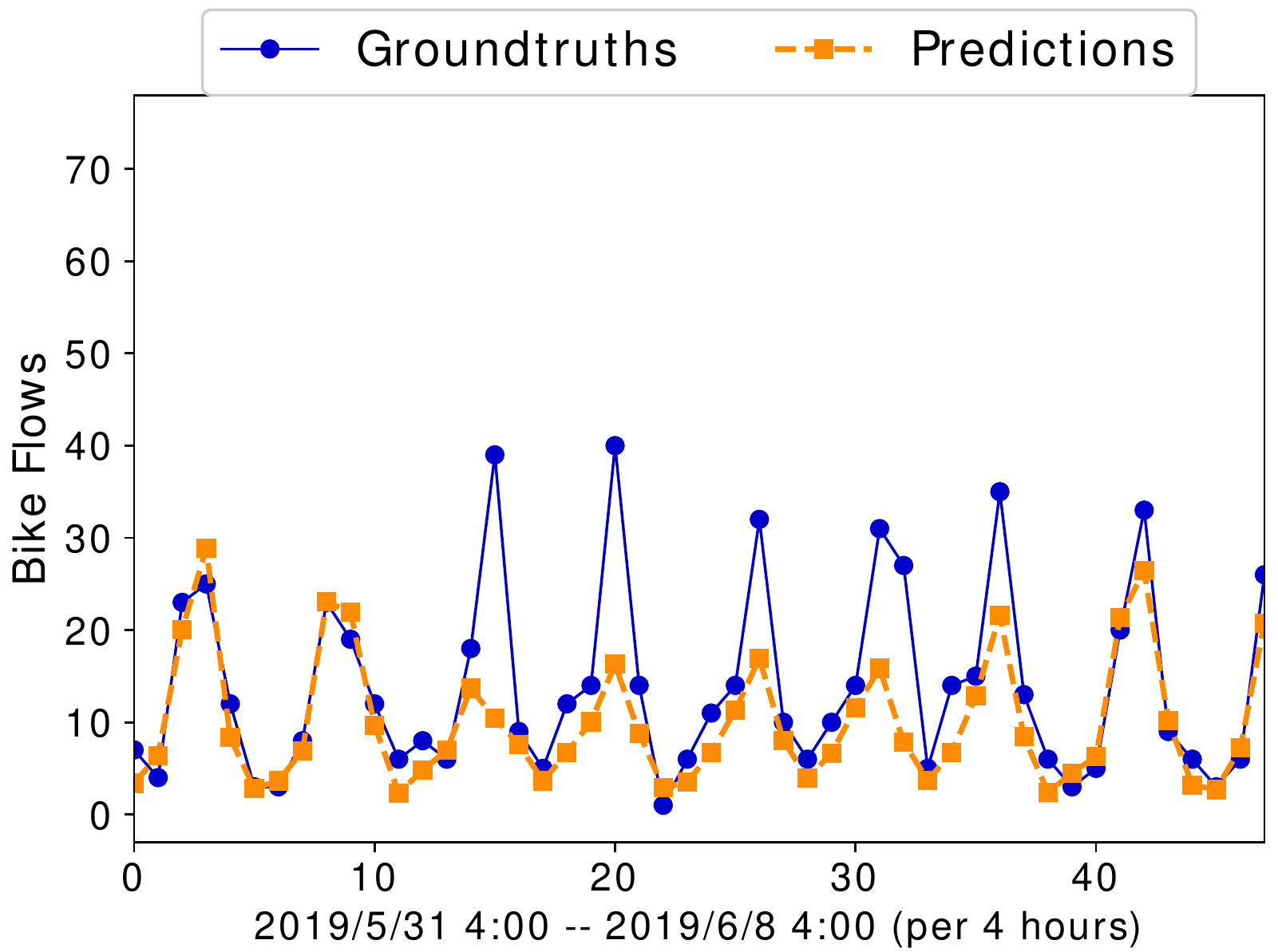}}
 	\subfigure[]{
 		\includegraphics[width=0.48\linewidth]{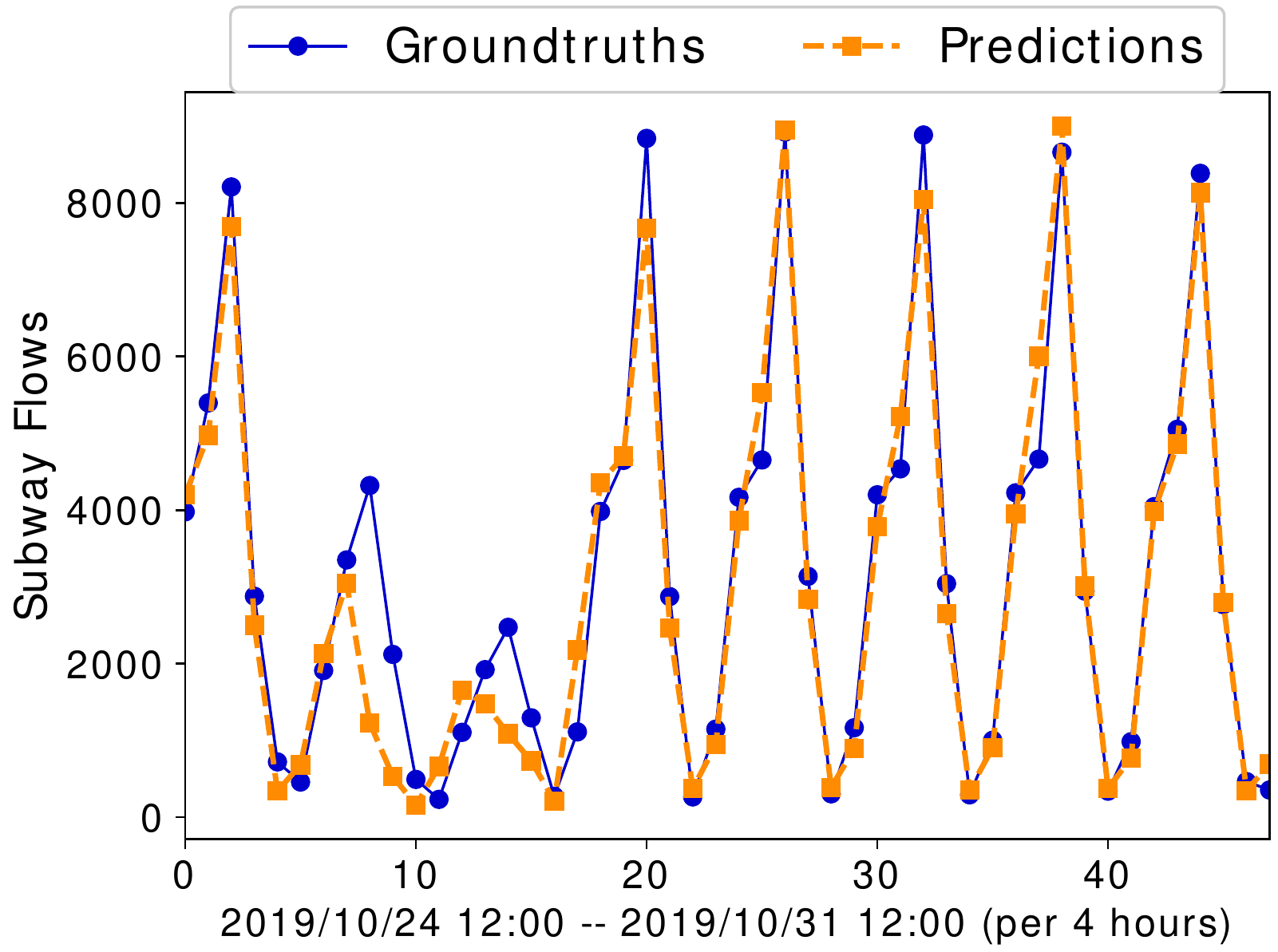}}
 	\subfigure[]{
 		\includegraphics[width=0.48\linewidth]{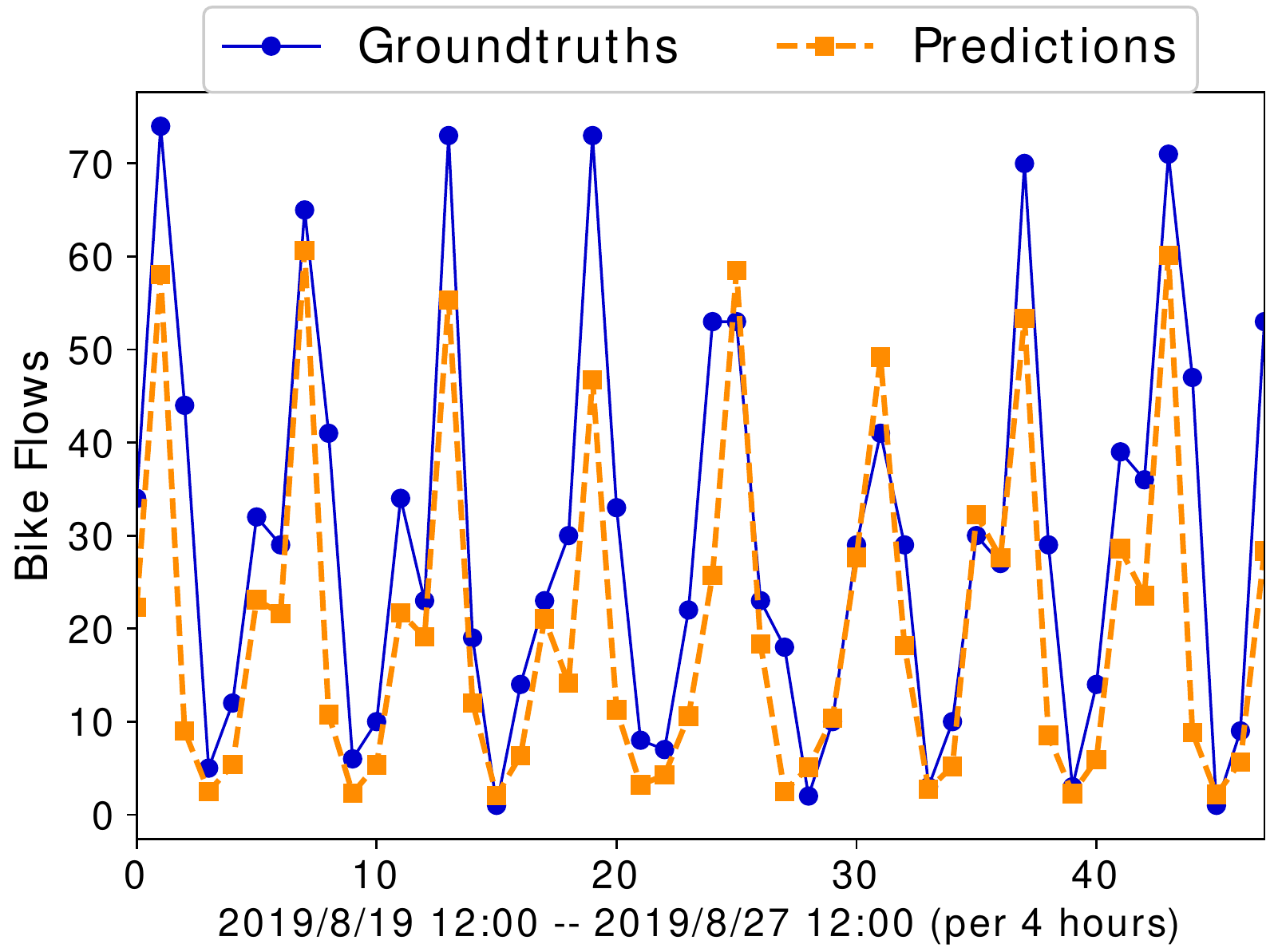}}
	\caption{(a) `Penn' subway station. (b) `21th St-49th Ave' bike station. (c) `40 Street - Lowery St' subway station. (d) `24th St-41st Ave' bike station.}
	\label{fig:casestudy}
\end{figure}

We display some result insights of four sampled stations. The selected timeslots are evenly distributed among the testing set. We can observe from Fig. \ref{fig:casestudy}: (1) `Penn' subway station is downtown of Manhattan and is around with abundant multi-mode crowd flows. Our method fits the curve well but may overestimate the potential flows. (2) In contrast, the bike station in (b) is sited just next to a large railway area without adequate crowd flows and POIs nearby. Our method tends to underestimate the potential flows. (3) the stations of (c)(d) are located in ordinary residential areas with regular crowd travel demands, and our method can model the diversion effects well. To summarise, even in the two extremes in (a)(b), our method can generate stable predictions of potential crowd flows for unseen new planned sites of different time slots in the long-range test sets.

\subsection{Ablation Study}

To verify the contribution of different components in MOHER to the performance gain, we further implement three simplified versions of our model to conduct ablation tests. The results of the degraded versions of MOHER on NYC Subway Turnstile Usage Dataset are shown in Table \ref{ablation}. We summarize the effects of three different components:

\begin{table}[th]
\centering
\caption{Effect of our method without each component on NYC Subway Turnstile Usage Data. Removing any component will result in a statistically significant error increase.}
\begin{tabular}{ccc}
\toprule[1pt]
Component                    & RMSE($\times 10^3$) & MAPE           \\ \hline
w/o cross-mode relations     & 1.776               & 1.499          \\
w/o functional similarity    & 1.689               & 1.247          \\
w/o differences modeling     & 1.861               & 1.346          \\
w/o regularization           & 1.676               & 1.471          \\
MOHER                        & \textbf{1.597}      & \textbf{1.124} \\ \bottomrule[0.5pt]
\end{tabular}
\label{ablation}
\end{table}

\subsubsection{Effect of cross-mode flow data} To investigate the effect of cross-mode dependency, we construct the graph with the geographical proximity and POI similarity only on the target mode (i.e., the subway). With the synthesis of the two metrics in Table \ref{ablation}, we can find that removing the cross-mode relations increases the error which justifies the importance of exploring the cross-mode flow data. Thus, the potential crowd flows of target new sites are implied in the various nearby transportation modes.

\subsubsection{Effect of modeling functional similarity} The regions with similar urban functions are considered to share similar flow patterns, so that we can infer the potential flow increments of new planned sites. The increased error demonstrates the importance of prior knowledge in functional similarity relations.

\subsubsection{Effect of modeling differences among multi-mode} Here we only model the correlations for cross-mode spatial dependencies extraction. The insufficient modeling of cross-mode relations results in a significant increase in error, which validates our intuition and observation.

\begin{figure}[t]
	\centering
	\includegraphics[width=0.49\linewidth]{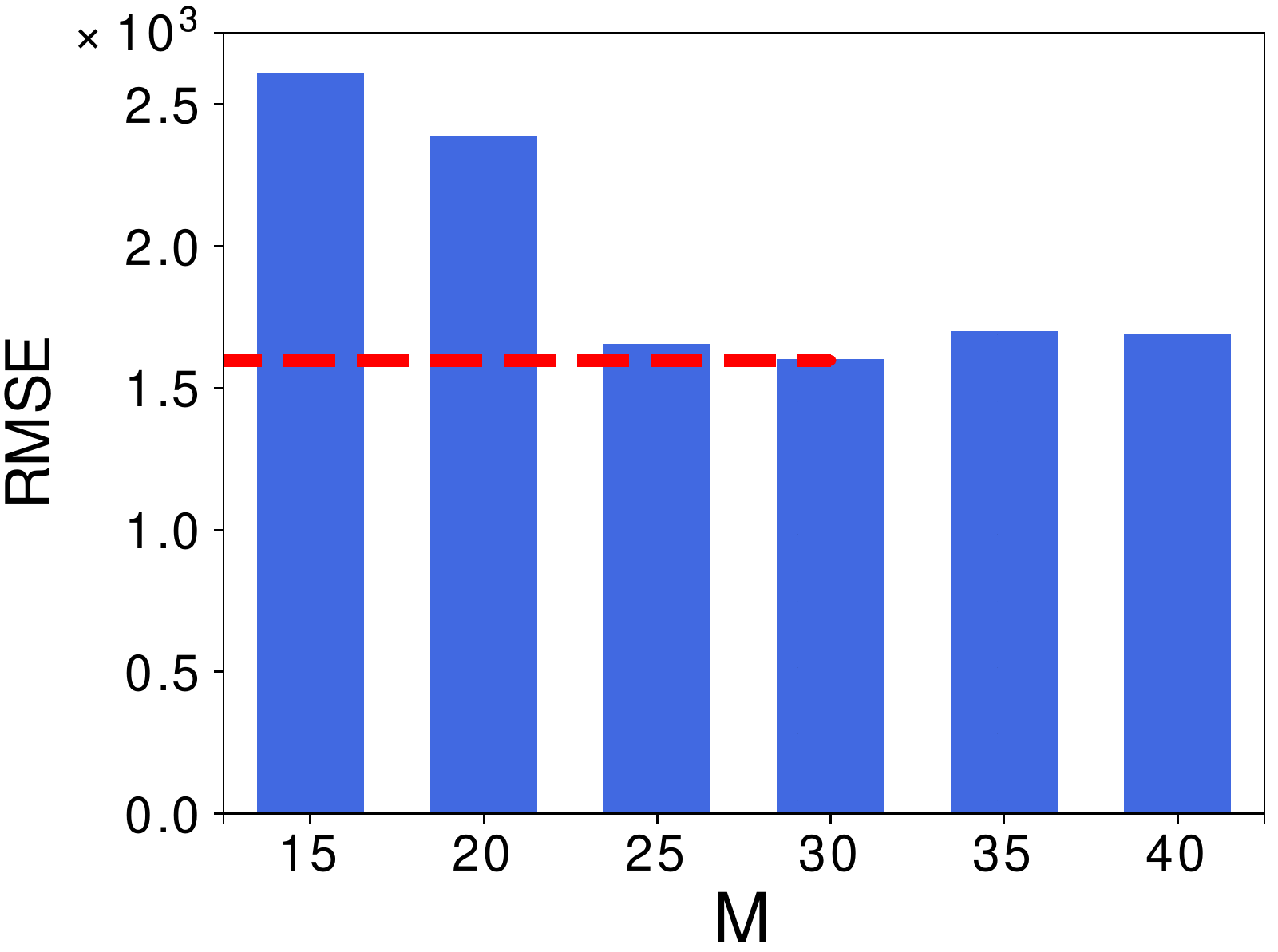}
	\includegraphics[width=0.49\linewidth]{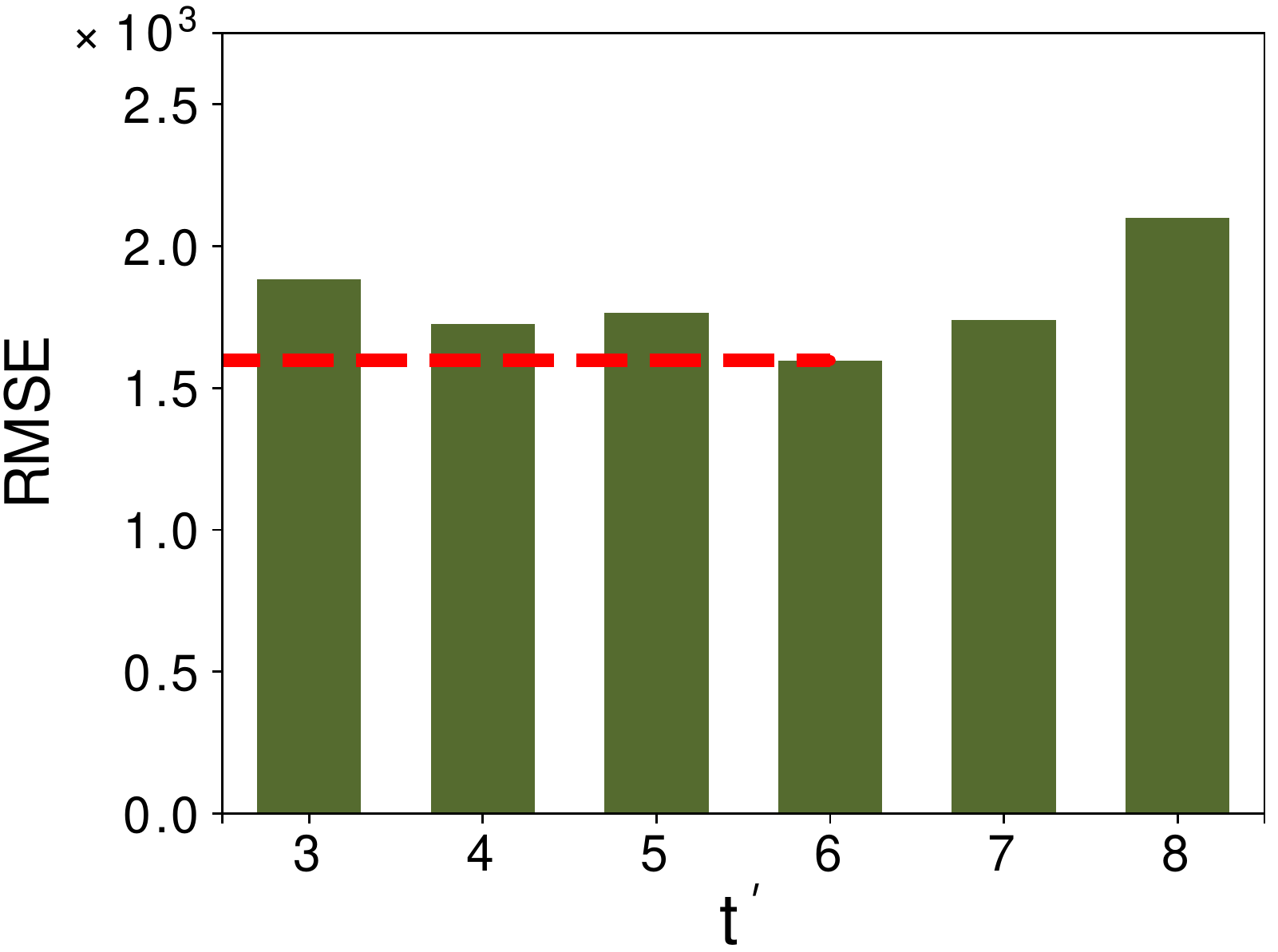}
	\caption{The Effect of Model Parameters}
	\label{fig:para}
\end{figure}

\subsubsection{Effect of matrix reconstruction regularization} To evaluate the importance of parameter regularization, we directly leverage the original FNN to model each type of heterogeneous relations. Because the high model complexity leads to overfitting, the degraded framework performs worse than MOHER.

\subsection{Effect of Model Parameters}
To study the effects of hyperparameters in our proposed framework, we present the performance on the Subway Dataset of two main hyperparameters for spatiotemporal prediction in Fig. \ref{fig:para}, i.e., the number of neighbor nodes in CFRGs $M$ and the length of time intervals $t'$. $M$ is the hyperparameter to control the boundary of modeling spatial dependency, and $t'$ controls the range of modeling temporal dependency. We observe that both the errors first decrease and then increase with the increase of $M$ and $t'$, because too small $M$ or $t'$ will both cause the lack of essential information for prediction. Meanwhile, larger $M$ will bring some unnecessary noise, and larger $t'$ may exceed the knowledge capacity of the hidden state in LSTM.

\section{Related Work}
How to mine the spatiotemporal information has become a long-standing problem for flow prediction \cite{lin2019deepstn, yao2019revisiting}. 
Some early works \cite{pan2019urban, wang2019origin} used geographical distance to model the relationship. 
Recently, there have been some researches \cite{li2018diffusion, zheng2020gman} on mining the complex relations of traffic flows. In particular, multiple heterogeneous relations have been explored as prior-knowledge for prediction \cite{chai2018bike, geng2019spatiotemporal, sun2020predicting, mgcgru}. 
\cite{bumingyigoutu} leveraged the graph embedding on heterogeneous spatial-temporal graphs for station-level demand prediction. 
\cite{ye2019co} combined the bike and taxi data for demand co-prediction.
However, there are only a few works towards unrecorded or potential flow prediction. Most of them are based on transductive tensor completion \cite{li2019tensor} to predict the historical missing data. In \cite{wu2020inductive}, the authors learned an inductive GNN for flows, which indicates the effect of GNN for unrecorded flow prediction. Moreover, \cite{aaai2020} intends to predict the potential passenger flows of planned subway stations based on the flow completion problem. Nevertheless, this method ignores the cross-mode crowd flows, thus having a diversion effect.


\section{Conclusion}
In this paper, we proposed MOHER, an inductive potential crowd flow prediction framework which can naturally generalize to future new planned sites. We encoded the heterogeneous relations among the target and its cross-mode neighbor sites/regions by measuring the geographical proximity and the functional similarity to capture the flow diversion and potential crowd flows. For explicitly modeling cross-mode heterogeneous relations, we developed a novel cross-mode relational GCN to learn the correlations and the differences between multiple transportation modes. Experimental results on real-world datasets demonstrated that the proposed MOHER framework outperforms the compared state-of-the-art methods with satisfactory margins.

\subsubsection{Acknowledgments}
This research is supported in part by National Natural Science Foundation of China (Grant No.62072235, 61572253).

\bibliography{bibliopraphy}
\end{document}